\documentclass[letterpaper, 10pt, conference]{ieeeconf}

\usepackage{graphicx}
\usepackage{lmodern}

\newcommand{\bm}[1]{\mbox{{\boldmath $#1$}}}

\IEEEoverridecommandlockouts
\title{\LARGE \bf
Initial Experiments on Learning-Based Randomized Bin-Picking\\ Allowing Finger Contact with Neighboring Objects
}

\author{
Kensuke Harada, Weiwei Wan, Tokuo Tsuji, Kohei Kikuchi, Kazuyuki Nagata, and Hiromu Onda
\thanks{K.~Harada is with Graduate School of Engineering Science, Osaka University, Toyonaka, Japan
       {\tt\small harada@sys.es.osaka-u.ac.jp}}%
\thanks{K.~Harada, W.~Wan, K.~Nagata and H.~Onda are with Intelligent Systems Research Institute, 
	National Institute of Advanced Industrial Science and Technology (AIST), Tsukuba, Japan}
\thanks{T.~Tsuji is with the Faculty of Engineering, Kanazawa University, Kanazawa, Japan}
\thanks{K.~Kikuchi is with Assembly Technology Department, Toyota Motors Co., Ltd., 1 Toyota-cho, Toyota 471-8572, Japan}%
}

\begin{document}

\maketitle
\thispagestyle{empty}
\pagestyle{empty}

\begin{abstract}
This paper proposes a novel method for randomized bin-picking based on learning. When a two-fingered gripper tries to pick an object from the pile, a finger often contacts a neighboring object. Even if a finger contacts a neighboring object, the target object will be 
successfully picked depending on the configuration of neighboring objects. In our proposed method, we use the visual information on neighboring objects to train the discriminator. Corresponding to a grasping posture of an object, the discriminator predicts whether or not the pick will be successful even if a finger contacts a neighboring object. We examine two learning algorithms, the linear support vector machine (SVM) and the random forest (RF) approaches. By using both methods, we demonstrate that the picking success rate is significantly higher than with conventional methods without learning. 
\end{abstract}

\section{INTRODUCTION}

Randomized bin-picking refers to the problem of automatically picking 
an object that is randomly stored in a box. If randomized bin-picking is introduced in a production process, 
we do not need parts-feeding machines or human workers 
to arrange the objects to be picked by a robot. However, randomized bin-picking is not widely 
introduced in production processes, because its success rate 
is not high (typically 80--85\%) \cite{Domae2014}. Although there are various reasons for this low success rate, 
this research focuses on one of the major problems, which can be explained as follows. 

\begin{figure}[t]
	\centering
	\includegraphics[width=5.5cm]{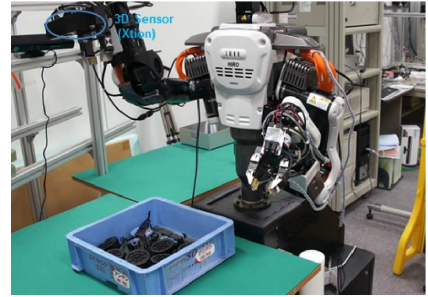}\\
	{\scriptsize (a) Overview of randomized bin-picking system}\\
	\includegraphics[width=4.5cm]{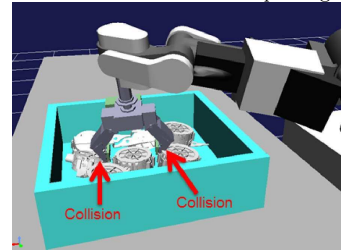}\\
        {\scriptsize (b) Failure of picking due to hand contact with neighboring objects}
	\caption{Overview of our bin-picking system \label{intro:RA-L2015}}
        \vspace{-5mm}
\end{figure}

Fig. \ref{intro:RA-L2015}(a) illustrates the randomized bin-picking of 
mechanical parts by using a two-fingered gripper. Fig. \ref{intro:RA-L2015}(b) 
shows a scene in which the gripper tries to grasp one of the objects. 
Because objects are placed close to each other, as shown in this figure, 
fingers may contact neighboring objects while the gripper approaches to 
the target object. 
In such cases, the success of the pick depends on the configuration of neighboring objects. 
For example, when a finger could slide in the gap between the target object and its neighbor, 
the pick tends to be successful even if a finger contacts the neighbor. 
On the other hand, the pick will not be successful if the finger first contacts a neighboring object 
which is not traversable while the gripper approaches to the target object. 
Hence, when picking an object from the pile, it is important to predict if the pick is successful before 
the hand actually grasps the target object. 

While there have been a number of studies on randomized bin-picking \cite{Domae2014,SII,Dupuis_08,Hujazi,Ghita,Kirkgaard,Fuchs,Zuo,icra13} 
and the grasp and manipulation planning of an object surrounded by many obstacles \cite{Dogar_RSS12,Berenson,Pas}, most approaches 
have attempted to realize a collision-free grasping posture. 
If we were to apply these methods to randomized bin-picking, 
the planner may sometimes cannot find a feasible grasping posture, 
because contact between a finger and a neighboring object cannot be avoided. 

To overcome the problem described above, we propose a novel approach for randomized bin-picking. 
Our method estimates of whether or not the pick will be successful based on the 
previous picking experience. 
Through a number of picking trials, a discriminator is trained 
based on relation between the visual information on neighboring objects 
and the result of pick. 
By using the trained discriminator, we consider predicting whether or not the pick will be 
successful even if a finger contacts a neighboring object. 
We examine two discriminators, one constructed using a linear support vector machine (SVM) 
and a second based on the random forest technique~\cite{Random}. 
Through some initial experiments, we show that, since we assumed large dimensional feature vector for the random-forest-based discriminator, it generates more accurate estimates. Experimental results show that, using both discriminators, 
the picking success rate becomes significantly higher than when using conventional methods. 

The remainder of this paper is organized as follows: after discussing some previous research 
in Section 2, Section 3 presents an overview of our bin-picking method. In Section 4, 
we describe the method used to learn the picking task. Section 5 explains the method for picking task execution. 
Section 6 contains an analysis of 
our experimental results, and Section 7 gives our conclusions. 

\section{RELATED WORK}

Randomized bin-picking has been studied by many researchers such as 
\cite{Domae2014,SII,Dupuis_08,Hujazi,Ghita,Kirkgaard,Fuchs,Zuo,icra13} while 
many of them focus on the image segmentation of randomly stacked objects 
and on the identification of their pose \cite{Hujazi,Ghita,Kirkgaard}. 
As for the research on randomized bin-picking focusing on the grasping capability of 
the hand, Dupuis et al.~\cite{Dupuis_08} applied the grasp planner MoveIt! to 
randomized bin-picking, whereas Domae et al.~\cite{Domae2014} proposed a method 
for picking an object based on the depth map of randomly stacked objects. 

For the grasp and manipulation planning of an object placed in a cluttered environment, 
Dogar et al.~\cite{Dogar_RSS12} proposed a method 
for pushing neighboring objects to obtain collision-free grasp of the target object. 
While several studies have examined learning approaches for 
grasping \cite{Curtis,Lenz,Pas,Ekvall}, 
most of them proposed methods for learning a grasping posture for a novel object. 
Pas et al.~\cite{Pas} developed a method for picking an object without 
using its geometrical model, instead learning the antipodal grasp using the SVN. 
Lenz et al.~\cite{Lenz} used deep learning to detect the appropriate grasp. 
However, in the research mentioned above, no one considers realizing 
the randomized bin-picking allowing the contact between a finger and neighboring objects. 

On the other hand, in the authors' previous work~\cite{icra13}, we proposed an approach 
for randomized bin-picking that allowed a finger to make contact with neighboring objects. 
However, the object shapes were limited to those that could be well-approximated by a set of 
cylinders. We believe that this study is the first attempt of randomized bin-picking taking 
the contact between a finger and a neighboring object with general shape into consideration. 

\section{BIN-PICKING OVERVIEW}

We first explain the method of randomized bin-picking used in our research. 
As shown in Fig. \ref{intro:RA-L2015}, let us consider the case in which the same objects are 
randomly stored in a box. By using a two-fingered gripper attached 
at the tip of a manipulator, we consider performing the randomize bin-picking. 

To pick an object from the pile, a 3D depth sensor (e.g., Xtion PRO) first 
captures a point cloud of randomly stored objects. Then, we segment the captured point cloud. 
In this research, we used a segmentation method based on the KD-tree prepared in the PCL (Point Cloud Library) \cite{pcl}. 
For each segment of point cloud which bounding-box size is similar to the bounding-box size of an object, we try to 
estimate the pose of an object. In this research, we used a two step algorithm for estimating the pose of an object: 
first roughly detecting the pose by using the 
CVFH (Clustered Viewpoint Feature Histogram) \cite{CVFH} and the CRH (Camera Roll Histogram) 
estimation, and then detecting the precise pose by using the ICP (Iterative Closest Point) algorithm, where all estimation methods are prepared in the PCL. 
Here, our estimation method usually estimates the poses of multiple objects. 

Then, we try to pick one of the objects which poses were detected. 
For a given geometrical model of an object, a set of grasping postures of the gripper for stably grasping 
an object is prepared in advance of starting the picking task, where each grasping posture is calculated 
by using a grasp planner such as \cite{Harada_icra08}. 
To pick an object, we consider selecting a grasping posture from multiple candidates of grasping postures 
among multiple objects. For each candidate of grasping posture, 
we solve IK to check the reachability of the robot. Then, for each reachable grasping 
posture, our proposed discriminator predicts whether or not the robot can successfully 
pick an object. We further select one of the grasping posture and perform the randomized bin-picking. 
This research defines that the pick is successful if the gripper grasps the target object, lifts it up, 
and places it out of the box. 

In the following section, we discuss how to train the discriminator, and how to execute the picking task by discriminating successful picks from failures. 

\section{Learning the Picking Task}

This section explains how to train the discriminator and to estimate whether the pick will succeed. 

\subsection{Swept Volume of Finger Motion}

\begin{figure}
	\centering
	\includegraphics[width=8cm]{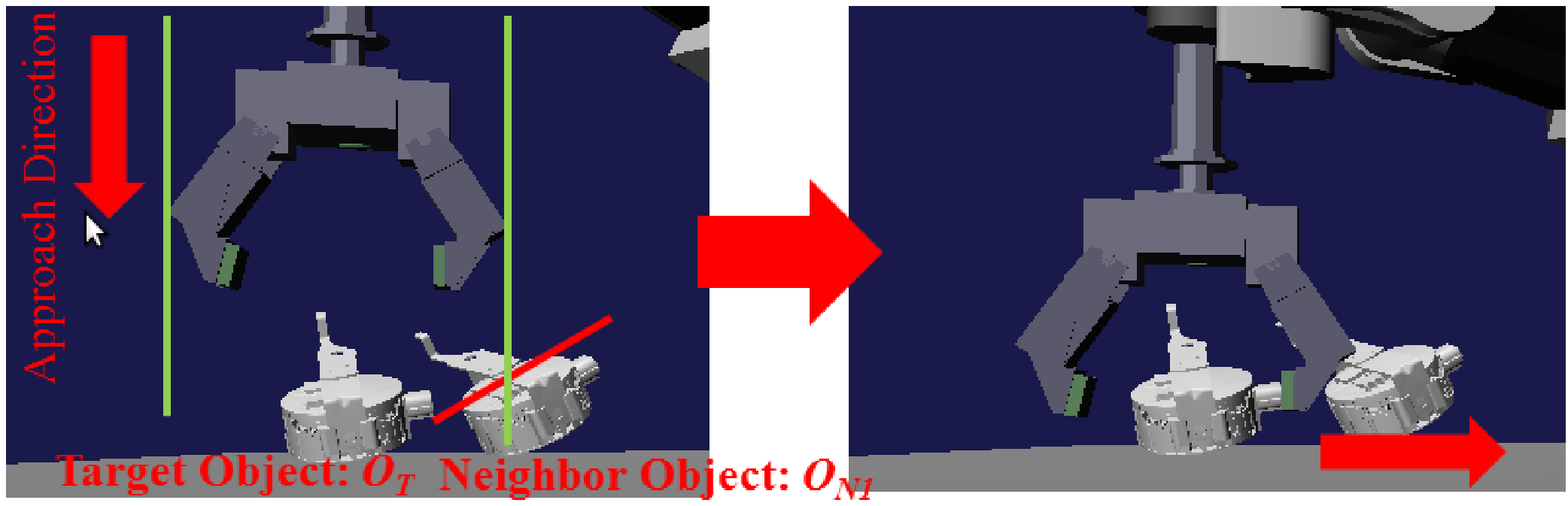}\\
	\vspace{1mm}
	{\scriptsize (a) Successful case of pick}\\
	\includegraphics[width=8cm]{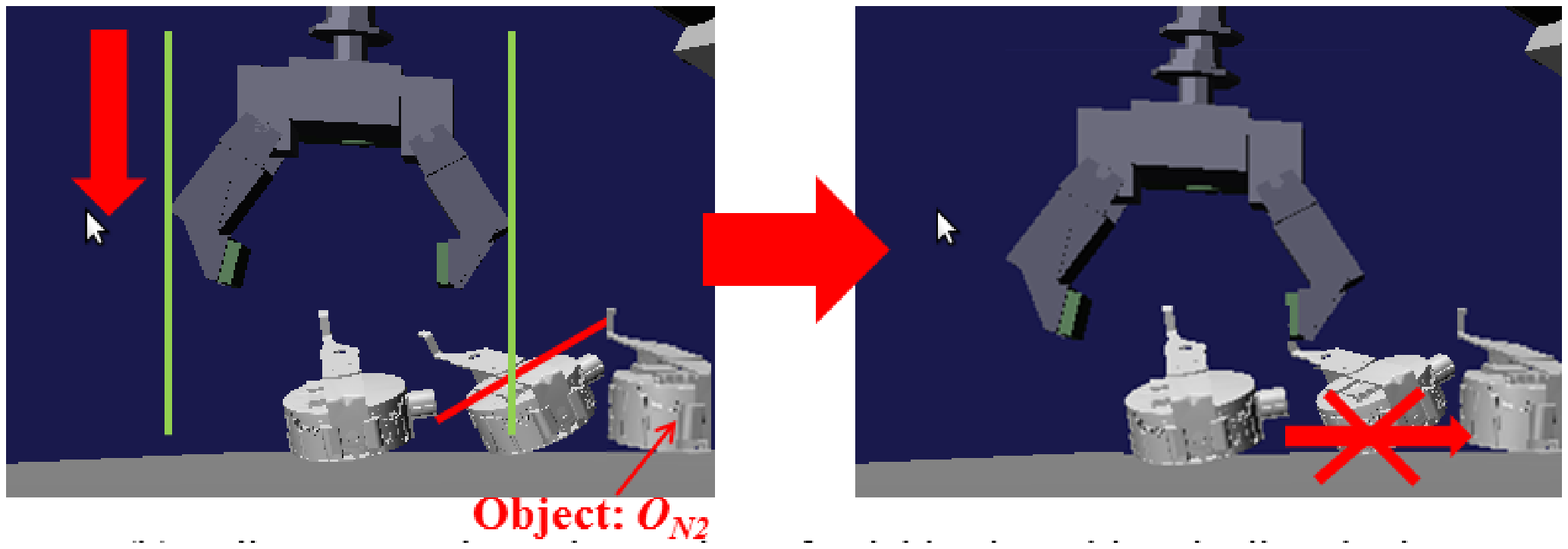}\\
	\vspace{1mm}
	{\scriptsize (b) Failure case where the motion of neighboring object is disturbed}\\
	\includegraphics[width=8cm]{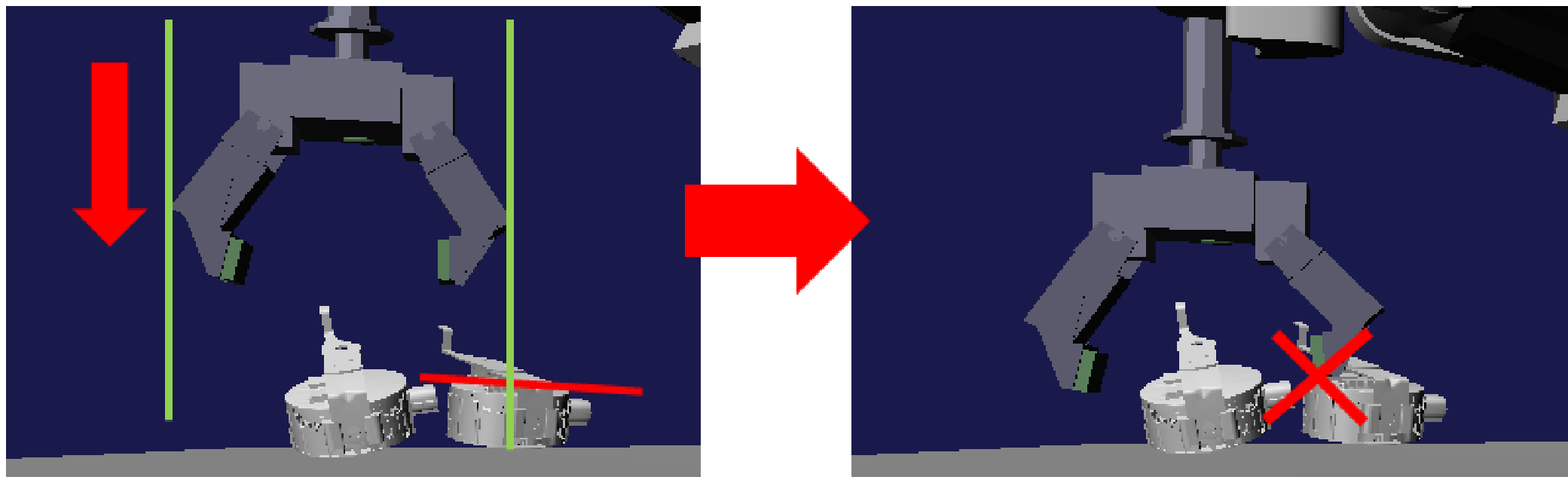}\\
	\vspace{1mm}
	{\scriptsize (c) Failure case where the contact angle between is different from (a)}
	\caption{Picking result depends on the contact angle between a finger and a neighboring object \label{fig:ContactAngle}}
        \vspace{-4mm}
\end{figure}

During a picking task, a two-fingered gripper first moves from the approach pose to the preshaping pose 
toward the approach direction (approach phase), and then fingers close to realize the grasping pose 
(grasp phase). 
Fig. \ref{fig:ContactAngle} shows some typical cases of contact between a finger and a neighboring object 
during the approach phase of a picking task. 
Fig. \ref{fig:ContactAngle} (a) shows a case where a finger contact the neighboring object $O_{N1}$ 
during the approach phase. In this case, since 
$O_{N1}$ is pushed by a finger and moves away from the target object $O_T$, a finger can be inserted into the gap between 
$O_T$ and $O_{N1}$. Hence, the robot will successfully pick the target object $O_T$. On the other hand, 
Fig. \ref{fig:ContactAngle} (b) shows a case where the object $O_{N2}$ is placed next to $O_{N1}$. In this case, 
depending on the travel distance of $O_{N1}$ moving away from $O_T$, the motion of $O_{N1}$ will be disturbed by $O_{N2}$. 
If the motion of $O_{N1}$ is disturbed, the pick may not be successful since a finger cannot be inserted 
into the gap between $O_T$ and $O_{N1}$. Although this observation shows that success of a pick depends on the 
configuration of all objects randomly stacked in a box, this research just focus on the region where a 
finger is expected to contact for simplicity. This region is sandwiched by two green lines shown 
in Fig. \ref{fig:ContactAngle}. This simplification is introduced for the initial trial of learning based 
randomized bin-picking by using a relatively low-dimensional feature vector to train the discriminator with 
relatively small number of samples. 
Based on the observation shown in Fig. \ref{fig:ContactAngle}, this simplification is justified when 
the motion of $O_{N1}$ is relatively small. 

\begin{figure}
	\centering
	\includegraphics[width=8cm]{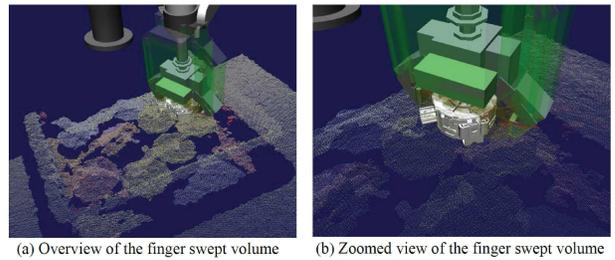}
	\caption{Swept volume of finger motion\label{fig:sweepvol}}
        \vspace{-0.3cm}
\end{figure}

\begin{figure}
	\centering
	\includegraphics[width=4.5cm]{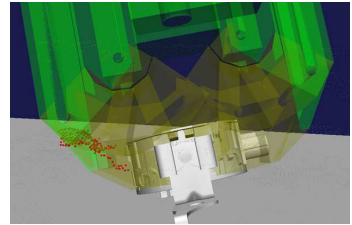}
	\caption{Point cloud included in the swept volume \label{fig:pointsVol}}
       \vspace{-0.4cm}
\end{figure}

As shown in Fig. \ref{fig:sweepvol}, we calculate the swept volume corresponding 
to the finger motion during the approach and grasp phases of the picking task, 
where we assumed that the fingers fully close at the grasping pose. 
Here, we note that this swept volume is calculated 
according to the planned motion of the fingers before the fingers actually move. 
This is because the swept volume is used to see the point-cloud distribution 
of neighboring objects which a finger is expected to contact. 
Hereafter, we name this swept volume as the {\it Swept Volume of Finger Motion} or simply 
the {\it Swept Volume}. 
Given a point cloud of stacked objects and a candidate grasping posture, 
we can obtain the distribution of point-cloud included in the swept volume as shown in Fig.\ref{fig:pointsVol}, 
where the point cloud included in the swept volume is denoted by the red dots. 
Here, from a point cloud included in the swept volume, 
we consider removing the points belonging to the target object by checking the distance between 
a point and the target object. 

\begin{figure}
	\centering
	\includegraphics[width=8cm]{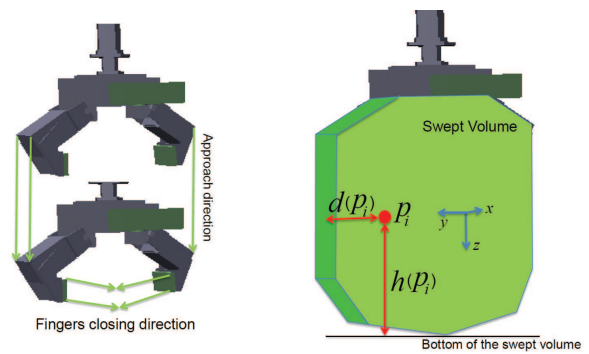}
	\caption{Definition of variables related to the finger swept volume \label{fig:heurist}}
\end{figure}

We will construct a discriminator predicting whether or not the pick is successful based on the 
distribution of point cloud included in the swept volume. 
For this purpose, let us assume a coordinate system attached to the swept volume, 
where the $z$ and $x$ axes denote the approach direction and the direction perpendicular to both 
the approach and the finger closing directions, as shown in Fig. \ref{fig:heurist}. 
Let $p_i$ $(i=1,\cdots,n)$ be the $i$-th point of the cloud included in the swept 
volume. As defined in Fig.~\ref{fig:heurist}, 
let $d(p_i)$ be the minimum distance between $p_i$ and the boundary of the finger swept volume in the 
$y$-direction, 
and let $h(p_i)$ be the distance between $p_i$ and the bottom of the finger swept volume in the 
$z$-direction. 

\subsection{Learning Algorithms}

By executing a series of picking experiment, we consider training a discriminator. 
To train a discriminator, we need the failure cases of pick as well as the successful cases. 
During the training phase, we did not use the discriminator predicting the success of the pick 
for the purpose of selecting a grasping posture from multiple candidates. 
Rather, we selected a grasping posture just by taking the grasp stability index \cite{icra14} 
into consideration. To train the discriminator, we use the information on the point-cloud distribution included in the 
swept volume and the information on whether or not the pick succeeded. 
We use the following two learning algorithms explained in the following:

\subsubsection{Linear SVM}

When picking the target object from the pile, 
the success of the pick depends on the configuration of the neighboring objects. 
If the finger could slide in the gap between the target object and its neighbor during the 
approach phase, the pick tends to be successful even if the finger contacts the neighbor. 
On the other hand, the pick will not be successful if the finger first contacts the neighbor 
and cannot slide in the gap between the target object and its neighbor. 
From this observation, we can assume the following two heuristic rules with respect to the distribution 
of point cloud included in the swept volume, i.e., the pick tends to be successful when the points are 
distributed at the edge of the swept volume, and the pick tends to be successful when the number of points 
included in the swept volume is small. The learning algorithm based on the linear SVM is constructed upon 
these two heuristic rules. 
By using the coordinate system fixed to the swept volume, $h(p_i)=0$ or $p(p_i)=0$ denote that 
the point $p_i$ $(i=1,\cdots,n)$ is on the edge of the swept volume. Also, $n=0$ denotes that 
there is no point included in the swept volume. For these cases, we can predict that the pick will 
be successful. Hence, the heuristic rules mentioned above can be expressed 
with respect to the coordinate system fixed to the swept volume that $h(p_i)$, $d(p_i)$ and $n$ are small 
$(i=1,\cdots,n)$. 

In the case of the linear SVM, we consider defining just a two dimensional feature vector based 
on this heuristic rule. Corresponding to the $j$-th trial of bin-picking, we define the following 
two dimensional feature vector: 

\begin{equation}
\bm{f}_{sj} = \left(\sum_{i=1}^n h(p_i), \  \sum_{i=1}^n d(p_i) \right)
\end{equation}

\noindent
Also, we define $r_j=1$ if the $j$-th trial of bin-picking was successful and 
-1 if the $j$-th trial was failed. 
By using a set of the training data ${\cal L} = \left\{ (\bm{f}_{sj}, r_j), j=1,\cdots,m \right\}$ 
obtained through a series of bin-picking, we consider training the discriminator. 


\subsubsection{Random Forest}

In the previous subsection, we defined a simple two dimensional feature vector just checking 
if a point cloud is distributed at the edge of the swept volume. However, we can imagine 
cases where such information on point-cloud distribution is not enough to predict the 
failure cases of pick. For example, the success of pick also depends on the contact angle 
between a finger and a neighboring object. 
In case shown in Fig. \ref{fig:ContactAngle} (c), even if a finger contact the neighboring object $O_{N1}$, $O_{N1}$ 
will not be pushed and will not move away from the target object $O_T$ due to the contact angle between 
a finger and $O_{N1}$. In this case, since a finger cannot be inserted into the gap between $O_T$ and $O_{N1}$, 
the pick will fail. 

Motivated by the need for more concrete information on the point cloud distribution, we applied to use a method for 
constructing a nonlinear discriminator named the Random Forest\cite{Random}. 
In the Random Forest method, we define the feature vector as follows. 
Corresponding to the distance to the boundary of the swept volume in the 
$y$ direction, we assume $b_y$ bins which width is $w_y$. 
Also, corresponding to the distance to the bottom of the finger swept volume in the 
$z$ direction, we assume $b_z$ bins which width is $w_z$. 
The point $p_i$ $(i=1,\cdots,n)$ is stored to the $j_y$-th $(\le b_y)$ and the $j_z$-th $(\le b_z)$ bins 
in the $y$ and the $z$ directions, respectively where their definitions are given by
\begin{eqnarray}
j_y &=& {\rm min}\left( \frac{d(p_i)}{w_y}, b_y \right), \\
j_z &=& {\rm min}\left( \frac{h(p_i)}{w_z}, b_z \right).
\end{eqnarray}

\begin{figure}
	\centering
	\includegraphics[width=7.5cm]{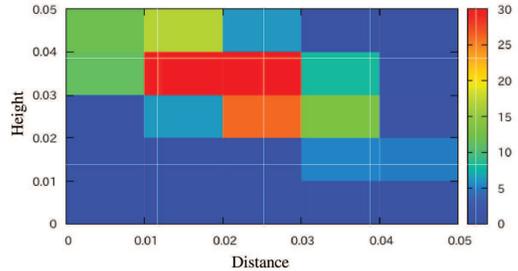}
	\caption{An example of the feature vector used in the random forest algorithm corresponding the point-cloud distribution shown in Fig. \ref{fig:pointsVol} where the number of points of each bin is shown. \label{fig:forest}}
        \vspace{-0.4cm}
\end{figure}

\noindent
After capturing a point cloud for the $j$-th trial of bin-picking, 
let us consider counting the number of points included in each bin. 
Let $\bm{f}_{rj}$ be the $b_y \cdot b_z$ dimensional 
feature vector where each element is the number of points included in each bin. 
Fig.\ref{fig:forest} shows the feature vector corresponding to the point cloud shown in 
Fig. \ref{fig:pointsVol} where we set $b_y=b_z=5$ and $w_y=w_z = 0.01${[m]}. 

To train the discriminator by using the training data 
${\cal L} = \left\{ (\bm{f}_{rj}, r_j), j=1,\cdots,m \right\}$  
obtained through a series of bin-picking experiment, 
the Random Forest algorithm first generates $N$ subsets of training data denoted by 
${\cal L}_k$ $(k=1,\cdots,N)$ by using the random sampling. For each subset, a decision tree is constructed. 
In case of the $k$-th decision tree $(k=1,\cdots,N)$, each node of the tree is a subset of ${\cal L}_k$. 
For example, let $\tilde{\cal L}_k$ be a subset of ${\cal L}_k$ forming a node of the $k$-th decision tree. 
To form its child nodes, we split $\tilde{\cal L}_k$ into $\tilde{\cal L}_k^L$ and $\tilde{\cal L}_k^R$ 
so as to minimize the Gini coefficient. We set the maximum depth of a decision tree to be $t_k$. 

To estimate whether the bin-picking will succeed, we first construct a feature vector and 
apply it to each decision tree. From each decision tree, we can obtain the success rate of the pick. By the mean of 
the success rate obtained from all the decision trees, we finally estimate whether the bin-picking will succeed. 
For the detailed algorithm of the random forest, refer \cite{Random}. 

\section{Picking Task Execution}

To pick an object from the pile, we have to determine a grasping posture of an object 
identified as a successful case of pick and having high grasp quality. 

For a given object, let ${\cal G}$ be a set of stable grasping configurations 
as briefly explained in Section 3 (called as the grasping configuration database). 
This set is generated before performing the picking task by using a grasp planner such as \cite{Harada_icra08} 
and is defined as 

\begin{equation}
{\cal G} = \left\{ (^o \bm{p}_i, ^o \bm{R}_i, \bm{\theta}_i, I_i), i=1,\cdots,d \right\}
\end{equation}

\noindent
where $^o\bm{p}_i$/$^o\bm{R}_i$ denote the position/orientation of the wrist with respect to the 
object coordinate system, and $\bm{\theta}_i$ and $I_i$ denote the finger joint angle vector 
and an index for evaluating the grasp stability such as \cite{icra14}, respectively. 

Once the poses of objects $\bm{p}_{oj}$/$\bm{R}_{oj}$ $(j=1,\cdots,e)$ are estimated by using a 3D depth sensor, 
we can obtain candidates of grasping configurations as follows:

\begin{equation}
{\cal G}_c = \left\{ (\bm{p}_{ij}, \bm{R}_{ij}, \bm{\theta}_i, I_i), i=1,\cdots,d, j=1,\cdots,e \right\} \label{eq:set}
\end{equation}

\noindent
where $\bm{p}_{ij}=\bm{p}_{oj}+\bm{R}_{oj}^o\bm{p}_i$ and $\bm{R}_{ij}=\bm{R}_{oj}^o\bm{R}_i$. 
Here, solving IK (inverse kinematics) for all the elements of eq.(\ref{eq:set}) 
may take a lot of time especially when the database size $d$ is large. 
We relax this problem by splitting eq.(\ref{eq:set}) into $f$ subsets according to the grasp quality index as follows:

\begin{eqnarray}
{\cal G}_{c1} &=& \left\{ (\bm{p}_{ij}, \bm{R}_{ij}, \bm{\theta}_i, I_i), i=1,\cdots,d, j=1,\cdots,e | \right. \nonumber \\
              && \left. I_i > t_1 \right\} \nonumber \\
              &\vdots& \nonumber \\
{\cal G}_{ck} &=& \left\{ (\bm{p}_{ij}, \bm{R}_{ij}, \bm{\theta}_i, I_i), i=1,\cdots,d, j=1,\cdots,e | \right. \nonumber \\
              && \left. I_i \le t_{k-1}, I_i > t_k \right\} \nonumber \\
              &\vdots& \nonumber \\
{\cal G}_{cf} &=& \left\{ (\bm{p}_{ij}, \bm{R}_{ij}, \bm{\theta}_i, I_i), i=1,\cdots,d, j=1,\cdots,e | \right. \nonumber \\
              && \left. I_i \le t_{f-1} \right\}  
\end{eqnarray}

From $k=1$ to $k=f$, we calculate the IK \footnote{To save the calculation time, we solved multiple IK problems in parallel. 
When solving IK, we also check the collision between a finger and the box. } 
and apply the discriminator to estimate the result of pick for all the elements of ${\cal G}_{ck}$. 
Then, if we can find at least one element of ${\cal G}_{ck}$ where the IK is solved and it is estimated as a successful case of pick, 
we select a grasping configuration from ${\cal G}_{ck}$. 
If there are multiple elements of ${\cal G}_{ck}$ where the IK is solved and it is estimated as a successful case of pick, 
we use an index function to select one of the elements. 
While we can assume several index functions 
such as the grasp quality measure $I_i$, the distance from the decision boundary of the linear SVM, and the probability obtained by 
using the random forest, this research used an intuitive rule of making $h(p_i)$, $d(p_i)$ and $n$ as small as possible 

\begin{equation}
I=- \sum_{i=1}^n (\alpha h(p_i) + \beta d(p_i)), 
\label{eq:heuris}
\end{equation}

\noindent
where $\alpha$ and $\beta$ are positive coefficients. We note that this index function is not used during 
the training phase of a discriminator by setting $\alpha = \beta =0$ for the purpose of collecting 
various failure cases. 

\section{EXPERIMENT}

We performed experiments on bin-picking. 
Overview of the robot system is shown in Fig. \ref{intro:RA-L2015}. 
We use the dual-arm manipulator HiroNX having two fingered 
gripper at the tip of each arm where each finger has two DOF. Since 
a 3D depth sensor (Xtion PRO) is attached at the wrist of the right arm, 
we used the left hand to pick an object. 

As shown in Fig. \ref{fig:segmentation}(a), we randomly placed nine objects in a box. 
We put nine objects close to each other such that the finger contacts 
a neighboring object when picking the target one. 

\subsection{Object Pose Estimation}

By using a 3D depth sensor attached at the wrist of the right arm, we capture the point 
cloud of the randomly stacked objects. 
Then, we consider segmenting the point cloud based on the distance between two points 
included in the point cloud as shown in Fig. \ref{fig:segmentation} (b) where 
each segment is expressed by different color. 
For the segment whose shape of its bounding box is similar to that 
of the object, we consider estimating its pose as shown in Fig. \ref{fig:segmentation} (c). 

When estimating the poses of objects by using the CVFH and ICP algorithms, we calculate the norm of 
the estimation error. For the purpose of removing the cases where the pick fails due to the 
estimation error of objects' pose, 
we did not calculate the candidates of grasping configurations 
of the objects with the norm of estimation error larger than the threshold where we set the threshold to be $0.007${[m]}. 
Within our trial, a finger always contacts a neighboring object when the pick fails. 
By using the multi-thread programming technique, we estimated the pose of multiple objects at the 
same time. By using a four-core 3GHz PC, it takes about 2 seconds to estimate the 
pose of eight objects with eight threads. 

\subsection{Learning}

To train the linear SVM, we prepared $m=50$ samples including 37 success and 13 failure cases. 
On the other hand, to train the random forest, we prepared $m=98$ samples including 
71 successful and 27 failure cases. Here, the random forest needs larger number of samples 
than the linear SVM to train the discriminator. 
As for the random forest, the number of data included in the subset ${\cal L}_k$ 
is set as 70$\%$ of that of the set ${\cal L}$. Also, we set $b_y=b_z=5$, $w_y=w_z = 0.01${[m]}, $N=200$ and $t_k=5$. 
While the number of samples needed for the random forest may change depending on the dimension of the feature vector, 
this used 25 dimensional feature vector. To explore the relation between the number of samples and the dimension 
of the feature vector, we need to perform a lot more experiments. 
Constructing experiment or simulation environment in which we can perform a number of picking tasks is 
considered to be our future research topic. 

\subsection{Picking Experiment}

To enable a comparison, we first examined the picking success rate without 
using the discriminator proposed 
in this research where we set $\alpha=\beta=0$ in the index function. 
In this case, the pick was successful in ten out of twenty trials. 

\begin{figure}
	\centering
	\includegraphics[width=7cm]{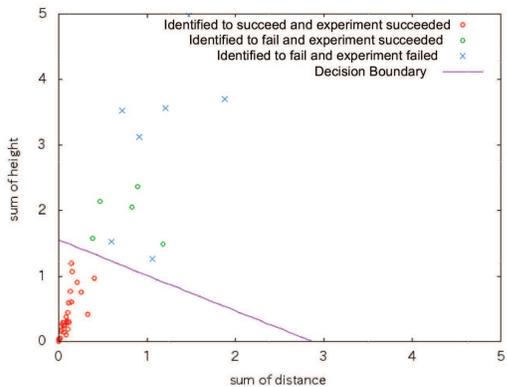}
	\caption{Results of picking experiment in which success/failure was identified by the linear SVM \label{fig:svm}}
\end{figure}

Table \ref{tab:svm} and Fig. \ref{fig:svm} show the results of picking experiment in which the success/failure cases were 
identified by using the linear SVM. Totally, the experiment succeeded for 40 out of 50 trials. 
The discriminator identified as successful for 33 out of 50 trials. Among 33 trials identified as successful, 
the experiment actually succeeded for all 33 trials. On the other hand, among 17 trials identified as failure, 
the experiment actually failed for 10 trials. 

\begin{table}
\caption{Results of picking experiment in which success/failure was identified by the linear SVM \label{tab:svm}}
\begin{tabular}{|l|c|c|} \hline 
 & {\scriptsize Identified as success} & {\scriptsize Identified as failure} \\ \hline
{\scriptsize Picking succeeded} & 33 & 7 \\ \hline 
{\scriptsize Picking failed}    & 0  & 10 \\ \hline
\end{tabular}
\end{table}

On the other hand, Table \ref{tab:rand} shows the result of using the random forest. 
Totally, the experiment succeeded for 42 out of 50 trials where the success rate is almost same as 
the experiment of using the linear SVM. 
The discriminator identified as successful for 39 out of 50 trials. Among 42 trials identified as successful, 
the experiment actually succeeded for 39 trials. On the other hand, among 11 trials identified as failure, 
the experiment actually failed for 5 trials. 

Let me consider the case where we stop picking an object if the discriminators predict the pick will fail. 
In this case, the success rate of pick will increase to 100 $\%$ for the cases of using the linear SVM and 92.9$\%$ for the 
cases of using the random forest which is the significant improvement from previous works \cite{Domae2014}. 

On the other hand, the rate of predicting successful cases is 66.0$\%$ for the cases of using the linear SVM and 
78.0$\%$ for the cases of using the random forest. This suggests that the random forest with higher dimension of 
the feature vector tends to give more accurate estimation. 

\begin{table}
\caption{Results of picking experiment in which the success/failure was identified by the random forest method \label{tab:rand}}
\begin{tabular}{|l|c|c|} \hline 
 & {\scriptsize Identified as success} & {\scriptsize Identified as failure} \\ \hline
{\scriptsize Picking succeeded} & 39 & 3 \\ \hline 
{\scriptsize Picking failed} & 3  & 5 \\ \hline
\end{tabular}
\end{table}

\begin{figure}
	\centering
	\includegraphics[width=8cm]{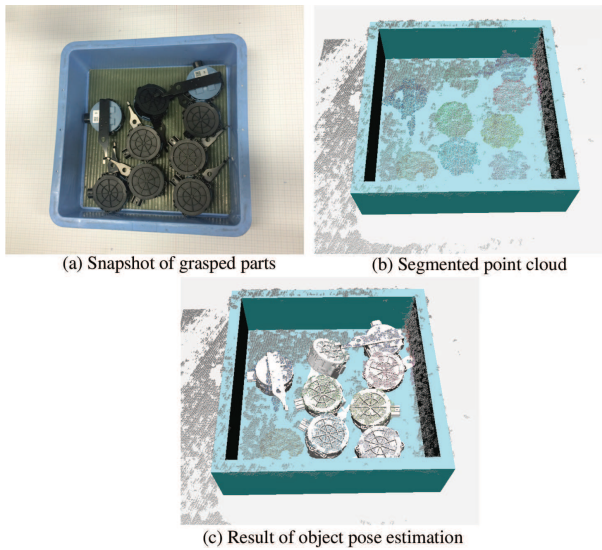}
	\caption{Estimation of objects' pose \label{fig:segmentation}}
        \vspace{-0.3cm}
\end{figure}

\begin{figure}
	\centering
	\includegraphics[width=6.5cm]{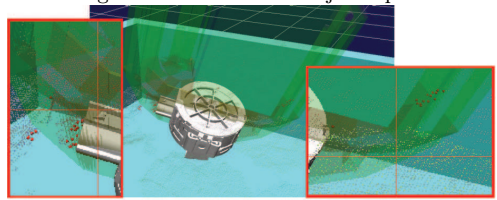}
	\vspace{-1mm}\\
	{\scriptsize (a) Swept volume of finger motion}
	\vspace{1mm}\\
	\includegraphics[width=5cm]{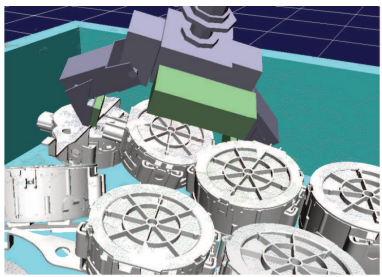}
	\vspace{-1mm}\\
	{\scriptsize (b) Calculated grasping posture}
	\vspace{-2mm}
	\caption{Result of grasping posture planning \label{fig:grasp}}
\end{figure}

\begin{figure}
	\centering
	\includegraphics[width=7cm]{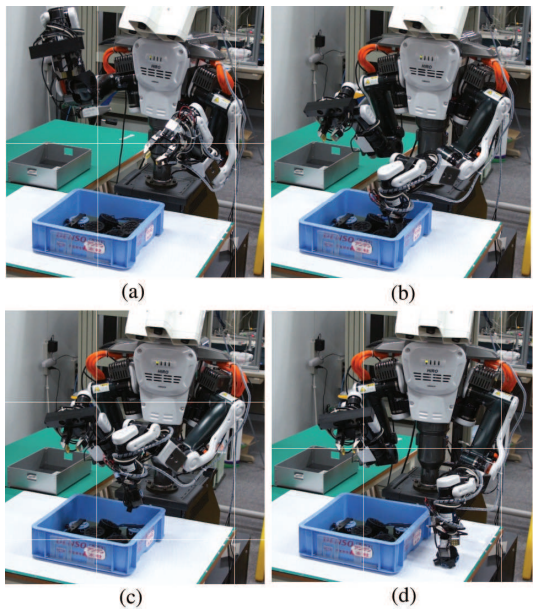}
	\caption{Overview of picking experiment \label{fig:experiment}}
        \vspace{-0.4cm}
\end{figure}

Figs. \ref{fig:segmentation}, \ref{fig:grasp} and \ref{fig:experiment} show 
a series of experimental result. 
For the object, we prepared the grasping configuration database which size 
is $d=172$. 
Fig. \ref{fig:segmentation} shows the result of 
pose estimation of objects where the pose of eight objects are estimated ($e=8$). 
We split the candidates of grasping configuration into three subsets ($f=3$) where 
we set $t_1$ and $t_2$ such that the size of each subset becomes as same as possible. 
We could find a feasible grasping configuration from ${\cal G}_{c1}$. Among 
$d  e / f \simeq 453$ candidates, 98 were IK solvable. Then, 50 out of 98 were 
identified to be a successful case of pick by using the random forest. 
Fig. \ref{fig:grasp} shows the selected grasping configuration and its finger swept volume. 
Here, in Fig. \ref{fig:grasp}(a), the red dot shows the point cloud included in the 
finger swept volume. Here, to cope with the sensor noise, we assumed a small margin (0.002{[m]}) to 
the size of the finger swept volume. Hence, in the figure, we can find some red dots 
out of the finger swept volume. Finally, Fig. \ref{fig:experiment} shows a series of 
experiment snapshot. In the experiment, although the finger contacts a neighboring object, 
the robot can successfully perform the picking task. 

\section{CONCLUSIONS}

In this paper, we proposed an approach on randomized bin-picking allowing contact between 
a finger and a neighboring object. 
By using the distribution of the point cloud 
obtained in the previous picking experiments, a discriminator is trained. 
The discriminator is used to predict whether or not the 
picking will be successfully performed even if a finger contacts a neighboring object. 
We used two learning algorithms, i.e., the linear SVM and the random forest. 
Through some elemental experimental results, we showed that the success rate of pick becomes 
significantly higher than that of the conventional methods. 

For the random bin-picking to be robust and reliable, we have to solve a lot of 
difficult problems. Among such difficult problems, the contribution of this work is that, 
by considering the point-cloud distribution of neighboring objects, we can 
predict the result of pick to some extent and can improve the success rate 
of picking task. The followings are some of the remaining problems: 
First, we will increase the number of samples and assume a large dimensional feature vector 
to train the discriminator. 
Second, performance of pick may further increase if we use time-series visual information 
to train the descriminator. 
Third, we will consider the effect of occlusion when picking an object from the pile. 
Fourth, we will remove heuristic rules from our learning framework. 
Fifth, extension of our proposed algorithm to more general multifingered hand is also considered to 
be our future research topic. 

\end{document}